\documentclass[11pt,a4paper]{article}
\usepackage[hyperref]{acl2021}
\usepackage{times}
\usepackage{float}
\usepackage{latexsym}
\usepackage{todonotes}
\usepackage[ruled,vlined]{algorithm2e}
\usepackage{caption}
\usepackage{subcaption}

\newcommand{\tabincell}[2]{\begin{tabular}{@{}#1@{}}#2\end{tabular}}

\usepackage{microtype}

\aclfinalcopy 
\def\aclpaperid{1293} 

\title{Positional Artefacts Propagate Through\\ Masked Language Model Embeddings}

\author{Ziyang Luo$^{1}$\thanks{\quad Work partly done during internship at NetEase Inc..}, Artur Kulmizev$^1$, Xiaoxi Mao$^2$\\
  $^1$  {\normalsize Department of Linguistics and Philology, Uppsala University, Sweden} \\
  $^2$ {\normalsize Fuxi AI Lab, NetEase Inc., Hangzhou, China} \\
  \texttt{\normalsize Ziyang.Luo.9588@student.uu.se, artur.kulmizev@lingfil.uu.se}\\
  \texttt{\normalsize maoxiaoxi@corp.netease.com}}

\date{}

\begin{document}
\maketitle
\begin{abstract}
  In this work, we demonstrate that the contextualized word vectors derived from pretrained
  masked language model-based encoders share a common, perhaps undesirable
  pattern across layers. Namely, we find cases of persistent outlier neurons
  within BERT and RoBERTa's hidden state vectors that consistently bear the
  smallest or largest values in said vectors. In
  an attempt to investigate the source of this information, we introduce a
  neuron-level analysis method, which reveals that the outliers are closely related to
  information captured by positional embeddings. We also pre-train the RoBERTa-base models from scratch and find that the outliers disappear without using positional embeddings. These outliers, we find, are the major
  cause of anisotropy of encoders' raw vector spaces, and \textit{clipping} them
  leads to increased similarity across vectors. We demonstrate this in practice
  by showing that clipped vectors can more accurately distinguish word senses, as well as lead to better sentence embeddings when mean pooling. In three supervised tasks, we find that \textit{clipping} does not affect the performance.
\end{abstract}

\section{Introduction}

A major area of NLP research in the deep learning era has concerned the representation of words in low-dimensional, continuous vector spaces. Traditional methods for achieving this have included word embedding models such as Word2Vec \citep{mikolov2013efficient}, GloVe \citep{pennington-etal-2014-glove}, and FastText
\citep{bojanowski-etal-2017-enriching}. However, though influential, such
approaches all share a uniform pitfall in assigning a single, static vector to a
word type. Given that the vast majority of words are polysemous \citep{KLEIN2001259}, static word embeddings cannot possibly represent a word's changing meaning in context.

In recent years, deep language models, like ELMo \citep{peters-etal-2018-deep}, BERT \citep{devlin-etal-2019-bert}
and RoBERTa \citep{liu2019roberta}, have achieved great success across many NLP
tasks. Such models introduce a new type of word vectors, deemed the
\textit{contextualized} variety, where the representation is computed with
respect to the context of the target word. Since these vectors are sensitive to
context, they can better address the polysemy problem that hinders
traditional word embeddings. Indeed, studies have shown that replacing
static embeddings (e.g. word2vec) with contextualized ones (e.g. BERT) can
benefit many NLP tasks, including constituency parsing
\citep{kitaev-klein-2018-constituency}, coreference resolution
\citep{joshi-etal-2019-bert} and machine translation
\citep{liu2020multilingual}.

However, despite the major success in deploying these representations across
linguistic tasks, there remains little understanding about information embedded
in contextualized vectors and the mechanisms that generate them. Indeed, an
entire research area central to this core issue --- the interpretability of
neural NLP models --- has recently emerged
\citep{ws-2018-2018,ws-2019-2019-acl,blackboxnlp-2020-blackboxnlp}. A key theme
in this line of work has been the use of linear probes in investigating the
linguistic properties of contextualized vectors \citep{tenney2018what,hewitt-manning-2019-structural}.
Such studies, among many others, show that
contextualization is an important factor that sets these embeddings apart from
static ones, the latter of which are unreliable in extracting features central
to context or linguistic hierarchy.
Nonetheless, much of this work likewise fails to engage with the raw vector
spaces of language models, preferring instead to focus its analysis on the
transformed vectors. Indeed, the fraction of work that has done the former has shed some curious insights: that untransformed BERT sentence representations still lag behind word embeddings across a variety of semantic benchmarks \citep{reimers-gurevych-2019-sentence} and that the vector spaces of language models are explicitly anisotropic \citep{ethayarajh-2019-contextual,li2020sentence}. Certainly, an awareness of the patterns inherent to models' untransformed vector spaces --- even if shallow --- can only benefit the transformation-based analyses outlined above. 

In this work, we shed light on a persistent pattern that can be observed for
contextualized vectors produced by BERT and RoBERTa. Namely, we show that,
across all layers, select neurons in BERT and RoBERTa consistently bear extremely large values. We observe this pattern across vectors for all words in several datasets, demonstrating that these singleton dimensions serve as
major outliers to the distributions of neuron values in both encoders'
representational spaces. With this insight in mind, the contributions of our
work are as follows:

\begin{enumerate}
\item We introduce a neuron-level method for analyzing the origin of a model's
  outliers. Using this, we show that they are closely related to positional
  information.
\item In investigating the effects of \textit{clipping} the outliers (zeroing-out),
  we show that the degree of anisotropy in the vector space diminishes significantly.
\item We show that after \textit{clipping} the outliers, the BERT representations can
  better distinguish between a word's potential senses in the word-in-context
  (WiC) dataset \citep{pilehvar-camacho-collados-2019-wic}, as well as lead to better sentence embeddings when mean pooling. 
 \end{enumerate}

\section{Finding outliers}\label{sec:catch}
\label{sec:catching}

In this section, we demonstrate the existence of large-valued vector dimensions across nearly all tokens encoded by BERT and RoBERTa. To illustrate these patterns, we employ two well-known datasets --- SST-2 \citep{socher-etal-2013-recursive} and QQP\footnote{\url{https://www.quora.com/q/quoradata/First-Quora-Dataset-Release-Question-Pairs}}. SST-2 (60.7k sentences) is a widely-employed sentiment analysis dataset of movie reviews, while QQP (727.7k sentences) is a semantic textual similarity dataset of Quora questions, which collects questions across many topics. We choose these datasets in order to account for a reasonably wide distributions of domains and topics, but note that any dataset would illustrate our findings well. We randomly sample 10k sentences from the training sets of both SST-2 and QQP, tokenize them, and encode them via BERT-base and RoBERTa-base. All models are downloaded from the Huggingface Transformers Library \citep{wolf-etal-2020-transformers}, though we replicated our results for BERT by loading the provided model weights via our own loaders.


When discounting the input embedding layers of each model, we are left with 3.68M and 3.59M contextualized token embeddings for BERT-base and RoBERTa-base, respectively. In order to illustrate the outlier patterns, we average all subword vectors for each layer of each model.



\begin{figure}
    \centering
    \includegraphics[height=5.4cm]{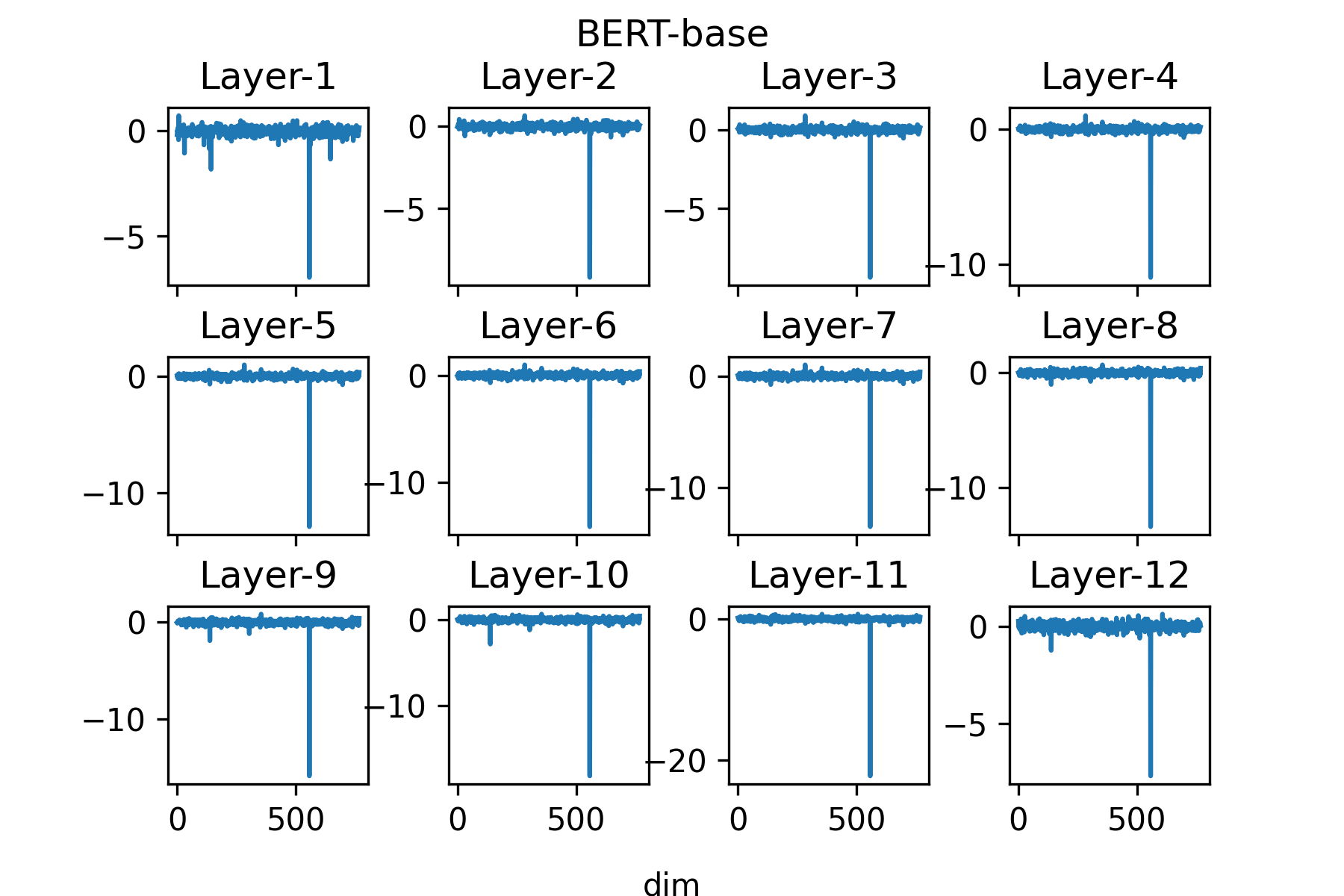}
    \caption{Average vectors for each layer of BERT-base.}
    \label{fig:bert-base-tail}
\end{figure}

\begin{figure}
    \centering
    \includegraphics[height=5.4cm]{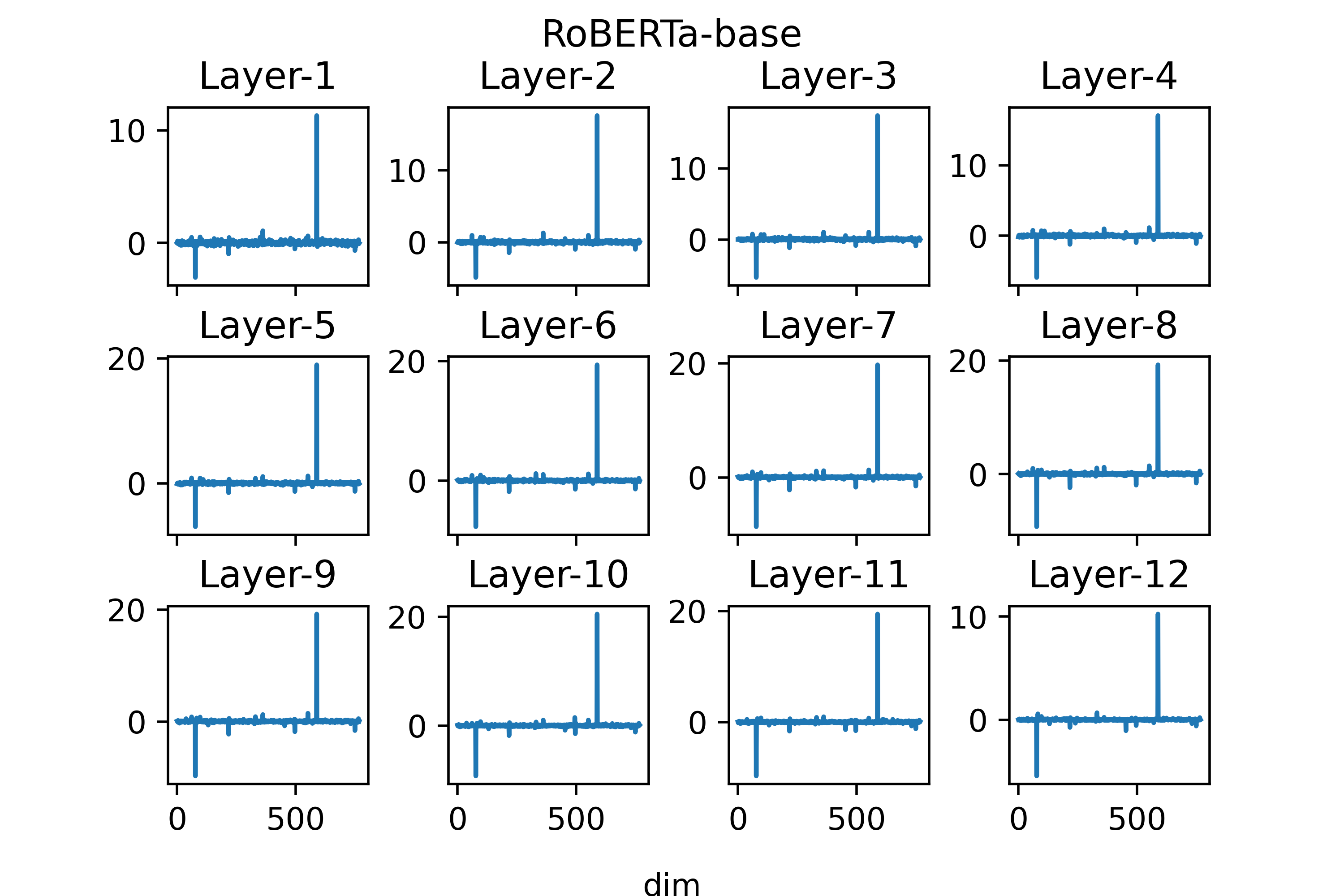}
    \caption{Average vectors for each layer of RoBERTa-base.}
    \label{fig:RoBERTa-base-tail}
\end{figure}




In examining BERT-base, we find that the minimum value of 96.60\% of vectors lies in the $557^{th}$ dimension. Figure \ref{fig:bert-base-tail} displays the averaged subword vectors for each layer of BERT-base, corroborating that these patterns exist across all layers. For RoBERTa-base, we likewise find that the maximum value of all vectors is the $588^{th}$ element. Interestingly, the \textit{minimum} element of 88.19\% of vectors in RoBERTa-base is the $77^{th}$ element, implying that RoBERTa has two such outliers. Figure \ref{fig:RoBERTa-base-tail} displays the average vectors for each layer of RoBERTa-base.

Our observations here reveal a curious pattern that is present in the base versions of BERT and RoBERTa. We also corroborate the same findings for the large and distilled \citep{sanh2020distilbert} variants of these architectures, which can be found in the Appendix \ref{app:outlier-distilled-large}. Indeed, it would be difficult to reach any sort of conclusion about the representational geometry of such models without understanding the outliers' origin(s). 

\section{Where do outliers come from?}\label{sec:where}

In this section, we attempt to trace the source of the outlier dimensions in BERT-base and RoBERTa-base (henceforth BERT and RoBERTa). Similarly to the previous section, we can corroborate the results of the experiments described here (as well as in the remainder of the paper) for the large and distilled varieties of each respective architecture. Thus, for reasons of brevity, we focus our forthcoming analyses on the base versions of BERT and RoBERTa and include results for the remaining models in the Appendix \ref{app:analysis-distilled-large} for the interested reader. 

\begin{figure}
    \centering
    \includegraphics[height=5cm]{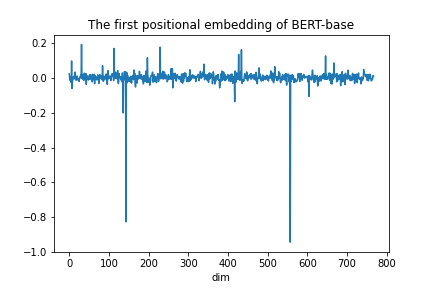}
    \caption{The first positional embedding of BERT-base.}
    \label{fig:bert-pos}
\end{figure}

\begin{figure}
    \centering
    \includegraphics[height=5cm]{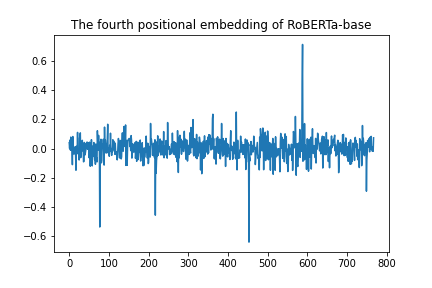}
    \caption{The fourth positional embedding of RoBERTa-base.}
    \label{fig:roberta-pos}
\end{figure}

In our per-layer analysis in §\ref{sec:catching}, we report that outlier dimensions exist across every layer in each model. Upon a closer look at the input layer (which features a vector sum of positional, segment, and token embeddings), we find that the same outliers also exist in positional embeddings. Figure \ref{fig:bert-pos} shows that the 1st positional embedding of BERT has two such dimensions, where the $557^{th}$ element is likewise the minimum. Interestingly, this pattern does not exist in other positional embeddings, nor in segment or token embeddings. Furthermore, Figure \ref{fig:roberta-pos} shows that the 4th positional embedding of RoBERTa has four outliers, which include the aforementioned $77^{th}$ and $588^{th}$ dimensions. We also find that, from the 4th position to the final position, the maximum element of $99.8\%$ positional embeddings is the $588^{th}$ element.

Digging deeper, we observe similar patterns in the Layer Normalization (LN, \citet{ba2016layer}) parameters of both models. Recall that LN has two learnable parameters --- \textit{gain} ($\gamma$) and \textit{bias} ($\beta$) --- both of which are 768-dimension vectors (in the case of the base models). These are designed as an affine transformation over dimension-wise normalized vectors in order to, like most normalization strategies, improve their expressive ability and to aid in optimization. Every layer of BERT and RoBERTa applies separate LNs post-attention and pre-output. For BERT, the $557^{th}$ element of the $\gamma$ vector is always among the top-6 largest values for the first ten layers' first LN. Specifically, it is the largest value in the first three layers. For RoBERTa, the $588^{th}$ element of the first LN's $\beta$ vector is always among the top-2 largest values for all layers --- it is largest in the first five layers. Furthermore, the $77^{th}$ element of the second LN's $\gamma$ are among the top-7 largest values from the second to the tenth layer.

It is reasonable to conclude that, after the vector normalization performed by LN, the outliers observed in the raw embeddings are lost. We hypothesize that these particular neurons are somehow important to the network, such that they retained after scaling the normalized vectors by the affine transformation involving $\gamma$ and $\beta$. Indeed, we observe that, in BERT, only the 1st position's embedding has such an outlier. However, it is subsequently observed in every layer and token after the first LN is applied. Since LayerNorm is trained globally and is not token specific, it happens to rescale every vector such that the positional information is retained. We corroborate this by observing that all vectors share the same $\gamma$. This effectively guarantees the presence of outliers in the 1st layer, which are then propagated upward by means of the Transformer's residual connection \citep{he2015deep}. Also, it is important to note that, in the case of BERT, the first position's embedding is directly tied to the requisite \texttt{[CLS]} token, which is prepended to all sequences as part of the MLM training objective. This has been recently noted to affect e.g. attention patterns, where much of the probability mass is distributed to this particular token alone, despite it bearing the smallest norm among all other vectors in a given layer and head \citep{kobayashi-etal-2020-attention}.\\

\noindent\textbf{Neuron-level analysis}\indent In order to test the extent to which BERT and RoBERTa's outliers are related to positional information, we employ a probing technique inspired by \citet{durrani-etal-2020-analyzing}. First, we train a linear probe $W\in R^{M\times N}$ without bias to predict the position of a contextualized vector in a sentence. In \citet{durrani-etal-2020-analyzing}, the weights of the classifier are employed as a proxy for selecting the most relevant neurons to the prediction. In doing so, they assume that, the larger the absolute value of the weight, the more important the corresponding neuron. However, this method disregards the magnitudes of the values of neurons, as a large weights do not necessarily imply that the neuron has high contribution to the final classification result. For example, if the value of a neuron is close to zero, a large weight also leads to a small contribution. In order to address this issue, we define the contribution of the $i^{th}$ neuron as $c(i)=abs(w_i*v_i)$ for $i=1,2,3,...,n$, where $w_i$ is the $i^{th}$ weight and $v_i$ is the $i^{th}$ neuron in the contextualized word vector. We name $C=[c(1), c(2),...,c(n)]$ as a contribution vector. If a neuron has a high contribution, this means that this neuron is highly relevant to the final classification result.

We train, validate, and test our probe on the splits provided in the SST-2 dataset (as mentioned in §\ref{sec:catching}, we surmise that any dataset would be adequate for demonstrating this). The linear probe is a $768\times 300$ matrix, which we train separately for each layer. Since all SST-2 sentences are shorter than 300 tokens in length, we set $M=300$. We use a batch size of 128 and train for 10 epochs with a categorical cross-entropy loss, optimized by Adam \citep{kingma2017adam}.

\begin{figure*}
     \centering
     \begin{subfigure}[t]{0.32\textwidth}
        \includegraphics[height=3.8cm]{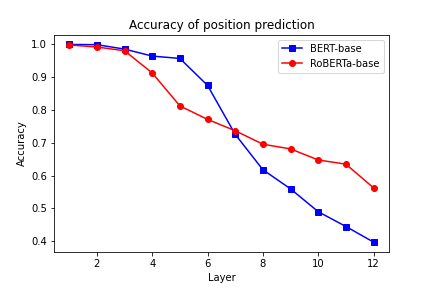}
        \centering
        \caption{Accuracy of position prediction.}
        \label{fig:acc}
     \end{subfigure}
     \hfill
    \begin{subfigure}[t]{0.32\textwidth}
        \includegraphics[height=3.8cm]{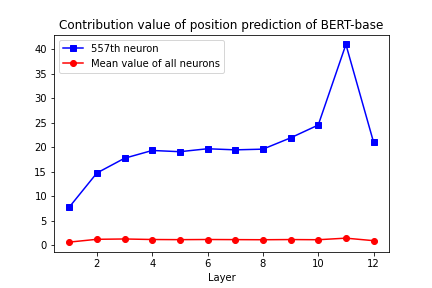}
        \centering
        \caption{The contribution value of BERT-base's outlier neuron on position prediction.}
        \label{fig:bert-contri}
    \end{subfigure}
    \hfill
    \begin{subfigure}[t]{0.32\textwidth}
        \includegraphics[height=3.8cm]{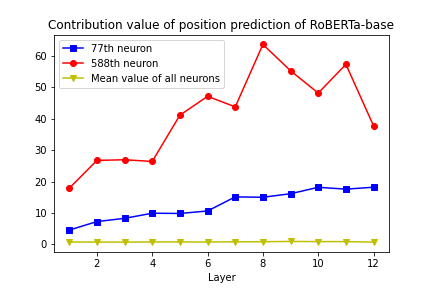}
        \centering
        \caption{The contribution value of RoBERTa-base's outlier neurons on position prediction.}
        \label{fig:roberta-contri}
    \end{subfigure}
\end{figure*}



Figure \ref{fig:acc} shows that, while it is possible to decode positional information from the lowest three layers with almost perfect accuracy, much of this information is gradually lost higher up in the model. Furthermore, it appears that the higher layers of RoBERTa contain more positional information than BERT. Looking at Figure \ref{fig:bert-contri}, we see that BERT's outlier neuron has a higher contribution in position prediction than the average contribution of all neurons. We also find that the contribution values of the same neuron are the highest in all layers. Combined with the aforementioned pattern of the first positional embedding, we can conclude that the $557^{th}$ neuron is related to positional information. Likewise, for RoBERTa, Figure \ref{fig:roberta-contri} shows that the $77^{th}$ and $588^{th}$ neurons have the highest contribution for position prediction. We also find that the contribution values of the $588^{th}$ neurons are always largest for all layers, which implies that these neurons are likewise related to positional information.\footnote{We also use heatmaps to show the contribution values in Appendix \ref{app:heatmap}.}\\


\noindent\textbf{Removing positional embeddings}\indent In order to isolate the relation between outlier neurons and positional information, we pre-train two RoBERTa-base models (with and without positional embeddings) from scratch using Fairseq \citep{ott-etal-2019-fairseq}. Our pre-training data is the English Wikipedia Corpus\footnote{We randomly select 158.4M sentences for training and 50k sentences for validation.}, where we train for 200k steps with a batch size of 256, optimized by Adam. All models share the same hyper-parameters, which are listed in the Appendix \ref{app:pre-train}. We use four NVIDIA A100 GPUs to pre-train each model, costing about 35 hours per model. 

We find that, without the help of positional embeddings, the validation perplexity of RoBERTa-base is very high at 354.0, which is in line with \citet{pmlr-v97-lee19d}'s observation that the self-attention mechanism of Transformer Encoder is order-invariant. In other words, the removal of PEs from RoBERTa-base makes it a bag-of-word model, whose outputs do not contain any positional information. In contrast, the perplexity of RoBERTa equipped with standard positional embeddings is much lower at 4.3, which is likewise expected.


In examining outlier neurons, we employ the same datasets detailed in §\ref{sec:catch}. For the RoBERTa-base model with PEs, we find that the maximum element of 82.56\% of all vectors is the $81^{st}$ dimension\footnote{Different initializations make our models have different outlier dimensions.}, similarly to our findings above. However, we do not observe the presence of such outlier neurons in the RoBERTa-base model without PEs, which indicates that the outlier neurons are tied directly to positional information. Similar to §\ref{sec:catch}, we display the averaged subword vectors for each layer of our models in Appendix \ref{app:our-model}, which also corroborate our results.

\section{Clipping the outliers}

In §\ref{sec:where}, we demonstrated that outlier neurons are related to positional information. In this section, we investigate the effects of zeroing out these dimensions in contextualized vectors, a process which we refer to as clipping. 

\subsection{Vector space geometry}\label{sec:geo}

\begin{figure*}
    \centering
    \includegraphics[width=\textwidth]{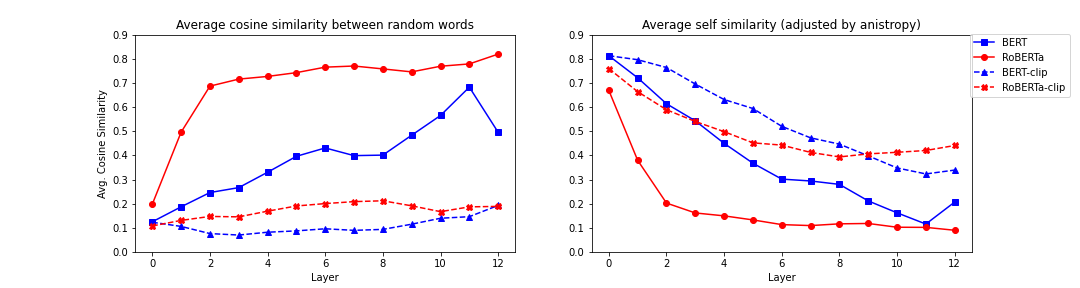}
    \caption{Left: anisotropy measurement of contextualized word vectors in BERT and RoBERTa before and after clipping the outlier dimensions. Right: self-similarity measurement of BERT and RoBERTa before and after clipping.}
    \label{fig:avg_cos_self_sim}
\end{figure*}

\noindent\textbf{Anisotropy}\indent\citet{ethayarajh-2019-contextual} observe that contextualized word vectors are anisotropic in all non-input layers, which means that the average cosine similarity between uniformly randomly sampled words is close to 1. To corroborate this finding, we randomly sample 2000 sentences from the SST-2 training set and create 1000 sentence-pairs. Then, we randomly select a token in each sentence, discarding all other tokens. This effectively sets the correspondence between the two sentences to two tokens instead. Following this, we compute the cosine similarity between these two tokens to measure the anisotropy of contextualized vectors. 

In the left plot of Figure \ref{fig:avg_cos_self_sim}, we can see that contextualized representations of BERT and RoBERTa are more anisotropic in higher layers. This is especially true for RoBERTa, where the average cosine similarity between random words is larger than 0.5 after the first non-input layer. This implies that the internal representations in BERT and RoBERTa occupy a narrow cone in the vector space.

Since outlier neurons tend to be valued higher or lower than all other contextualized vector dimensions, we hypothesize that they are the main culprit behind the degree of observed anisotropy. To verify our hypothesis, we clip BERT and RoBERTa's outliers by setting each neuron's value to zero. The left plot in Figure \ref{fig:avg_cos_self_sim} shows that, after clipping the outliers, their vector spaces become close to isotropic. \\

\noindent\textbf{Self-similarity} In addition to remarking upon the anisotropic characteristics of contextualized vector spaces, \citet{ethayarajh-2019-contextual} introduce several measures to gauge the extent of ``contextualization'' inherent models. One such metric is \textit{self-similarity}, which the authors employ to compare the similarity of a word's internal representations in different contexts. Given a word $w$ and $n$ different sentences $s_1,s_2,...,s_n$ which contain such word, $f_l^i(w)$ is the internal representation of $w$ in sentence $s_i$ in the $l^{th}$ layer. The average self-similarity of $w$ in the $l^{th}$ layer is then defined as:
\begin{equation}
    \mathrm{SelfSim}_l(w)=\frac{\sum_{i=1}^n\sum_{j=i+1}^n \cos\left(f_l^i(w),f_l^j(w)\right)}{n(n-1)}
\end{equation}
Intuitively, a self-similarity score of 1 indicates that no contextualization is being performed by the model (e.g. static word embeddings), while a score of 0 implies that representations for a given word are maximally different given various contexts. 

To investigate the effect of outlier neurons on a model's self-similarity, we sample 1000 different words from SST-2 training set, all of which appear at least in 10 different sentences. We then compute the average self-similarity of these words as contextualized by BERT and RoBERTa --- before and after clipping the outliers. To adjust for the effect of anisotropy, we subtract the self-similarity from each layer's anisotropy measurement, as in \citet{ethayarajh-2019-contextual}.

The right plot in Figure \ref{fig:avg_cos_self_sim} shows that, similarly to the findings in \citep{ethayarajh-2019-contextual}, a word's self-similarity is highest in the lower layers, but decreases in higher layers. Crucially, we also observe that, after clipping the outlier dimensions, the self-similarity increases, indicating that vectors become closer to each other in the contextualized space. This bears some impact on studies attempting to characterize the vector spaces of models like BERT and RoBERTa, as it is clearly possible to overstate the degree of ``contextualization'' without addressing the effect of positional artefacts. 

\subsection{Word sense}

Bearing in mind the findings of the previous section, we now turn to the question of word sense, as captured by contextualized embeddings. Suppose that we have a target word $w$, which appears in two sentences. $w$ has the same sense in these two sentences, but its contextualized representations are not identical due to the word appearing in (perhaps slightly) different contexts. In the previous few sections, we showed that outlier neurons are related to positional information and that clipping them can make a word's contextualized vectors more similar. Here, we hypothesize that clipping such dimensions can likewise aid in intrinsic semantic tasks, like differentiating senses of a word. 

To test our hypothesis, we analyze contextualized vectors using the word-in-context (WiC) dataset \citep{pilehvar-camacho-collados-2019-wic}, which is designed to identify the meaning of words in different contexts. WiC is a binary classification task, where, given a target word and two sentences which contain it, models must determine whether the word has the same meaning across the two sentences. 



In order to test how well we can identify differences in word senses using contextualized vectors, we compute the cosine similarity between contextualized vectors of target words across pairs of sentences, as they appear in the WiC dataset. If the similarity value is larger than a specified threshold, we assign the true label to the sentence pair; otherwise, we assign the false label. We use this method to compare the accuracy of BERT and RoBERTa on WiC before and after clipping the outliers. Since this method does not require any training, we test our models on the WiC training dataset.\footnote{The WiC test set does not provide labels and the size of validation set is too small (638 sentences pairs). We thus choose to use the training dataset (5428 sentences pairs).} We compare 9 different thresholds from 0.1 to 0.9, as well as a simple baseline model that assigns the true labels to all samples.

\begin{table}
    \centering
    \small
    \begin{tabular}{lccl}
        \hline
        Model & Layer & Threshold & Accuracy \\
        \hline
        \hline
        \textbf{Baseline} & - & - & 50.0\%\\
        \hline
        \hline
        \textbf{Before clipping}\\
        BERT & 7 & 0.7 & 67.5\%\\
        RoBERTa & 10 & 0.9 & 69.0\%\\
        \hline
        \hline
        \textbf{After clipping}\\
        BERT-clip & 10 & 0.5 & \textbf{68.4\%}\\
        RoBERTa-clip & 11 & 0.6 & \textbf{69.9\%}\\
        \hline
    \end{tabular}
    \caption{The best accuracy scores on WiC dataset. \textbf{Bold} indicates that the best result increases after clipping.}
    \label{tab:wic}
\end{table}

\begin{table*}[ht!]
    \centering
    \small
    \begin{tabular}{llllllll}
        \hline
        Dataset & STS-B & SICK-R & STS-12 & STS-13 & STS-14 & STS-15 & STS-16 \\
        \hline
        \hline
        \textbf{Baseline}\\
        Avg. GloVe & 58.02 & 53.76 & 55.14 & 70.66 & 59.73 & 68.25 & 63.66\\
        \hline
        \hline
        \textbf{Before clipping}\\
        BERT & 58.61(3) & 60.78(2) & 48.00(1) & 61.19(12) & 50.10(12) & 61.15(1) & 62.38(12)\\
        RoBERTa & 56.60(11) & 64.68(11) & 40.00(1) & 58.33(11) & 49.79(8) & 64.39(9) & 64.82(11)\\
        \hline
        \hline
        \textbf{After clipping}\\
        BERT-clip & \textbf{63.06(2)} & \textbf{61.74(2)} & \textbf{50.40(1)} & \textbf{61.44(1)} & \textbf{54.52(2)} & \textbf{67.00(2)} & \textbf{64.18(2)}\\
        RoBERTa-clip & \textbf{60.61(11)} & \textbf{64.82(11)} & \textbf{43.44(1)} & \textbf{59.72(11)}& \textbf{51.92(3)}& \textbf{66.15(3)} & \textbf{67.14(11)}\\
        \hline
    \end{tabular}
    \caption{Experimental results on semantic textual similarity, where the baselines results are published in \citet{reimers-gurevych-2019-sentence}. We show the best Spearman rank correlation between sentence embeddings' cosine similarity and the golden labels. The results are reported as $r\times 100$. The number in the parenthesis denotes that this result belongs to the specific layer. \textbf{Bold} indicates that the best result increases after clipping.}
    \label{tab:sts_results}
\end{table*}

Table \ref{tab:wic} shows that after clipping outliers, the best accuracy scores of BERT and RoBERTa increase about $1\%$.\footnote{The thresholds are different due to the fact that the cosine similarity is inflated in the presence of outlier neurons.} This indicates that these neurons are less related to word sense information and can be safely clipped for this particular task (if performed in an unsupervised fashion). 

\subsection{Sentence embedding}\label{sub:sent}

Venturing beyond the word-level, we also hypothesize that outlier clipping can lead to better sentence embeddings when relying on the cosine similarity metric. To test this, we follow \citet{reimers-gurevych-2019-sentence} in evaluating our models on 7 semantic textual similarity (STS) datasets, including the STS-B benchmark (STS-B) \citep{cer-etal-2017-semeval}, the SICK-Relatedness (SICK-R) dataset \citep{sick} and the STS tasks 2012-2016 \citep{agirre-etal-2012-semeval,agirre-etal-2013-sem,agirre-etal-2014-semeval,agirre-etal-2015-semeval,agirre-etal-2016-semeval}. Each sentence pair in these datasets is annotated with a relatedness score on a 5-point rating scale, as obtained from human judgments. We load each dataset using the SentEval toolkit \citep{conneau-kiela-2018-senteval}.  

Indeed, the most common approach for computing sentence embeddings from contextualized models is simply averaging all subword vectors that comprise a given sentence \citep{reimers-gurevych-2019-sentence}. We follow this method in obtaining embeddings for each pair of sentences in the aforementioned tasks, between which we compute the cosine similarity. Given a set of similarity and gold relatedness scores, we then calculate the Spearman rank correlation. As a comparison, we also consider averaged GloVe embeddings as our baseline.


Table \ref{tab:sts_results} shows that, after clipping the outliers, the best Spearman rank correlation scores for BERT and RoBERTa increase across all datasets, some by a large margin. This indicates that clipping the outlier neurons can lead to better sentence embeddings when mean pooling. However, like \citet{li-etal-2020-sentence}, we also notice that averaged GloVe embeddings still manage outperform both BERT and RoBERTa on all STS 2012-16 tasks. This implies that the post-clipping reduction in anisotropy is only a partial explanation for why contextualized, mean-pooled sentence embeddings still lag behind static word embeddings in capturing the semantics of a given sentence. 

\subsection{Supervised tasks}

In the previous sections, we analyzed the effects of clipping outlier neurons on various intrinsic semantic tasks. Here, we explore the effects of clipping in a supervised scenario, where we hypothesize that a model will learn to discard outlier information if it is not needed for a given task. We consider two binary classification tasks, SST-2 and IMDB \citep{maas-EtAl:2011:ACL-HLT2011}, and a multi-class classification task, SST-5, which is a 5-class version of SST-2. First, we freeze all the parameters of the pre-trained models and use the same method in §\ref{sub:sent} to get the sentence embedding of each sentence. Then, we train a simple linear classifier $W\in R^{768\times N}$ for each layer, where $N$ is the number of classes. We use different batch sizes for different tasks, 768 for SST-2, 128 for IMDB and 1536 for SST-5. Then we train for 10 epochs with a categorical cross-entropy loss, optimized by Adam. 

\begin{table}
    \centering
    \small
    \begin{tabular}{lccc}
        \hline
        Model & SST-2 & IMDB & SST-5\\
        \hline
        \hline
        \textbf{Before clipping}\\
        BERT & 85.9\%(12) & 86.8\%(10) & 46.2\%(10)\\
        RoBERTa & 88.4\%(8) & 91.5\%(9) & 46.9\%(7)\\
        \hline
        \hline
        \textbf{After clipping}\\
        BERT-clip & 85.4\%(12) & 86.4\%(10) & 46.1\%(12)\\
        RoBERTa-clip & 88.7\%(8) & 91.6\%(9) & 47.0\%(7)\\
        \hline
    \end{tabular}
    \caption{The best accuracy scores on different supervised tasks. The number in the parenthesis denotes that this result belongs to the specific layer.}
    \label{tab:sst-2}
\end{table}

Table \ref{tab:sst-2} shows that there is little difference in employing raw vs. clipped vectors in terms of task performance. This indicates that using vectors with clipped outliers does not drastically affect classifier accuracy when it comes to these common tasks. 

\section{Discussion}


The experiments detailed in the previous sections point to the dangers of relying on metrics like cosine similarity when making observations about models' representational spaces. This is particularly salient when the vectors being compared are taken off-the-shelf and their composition is not widely understood. Given the presence of model idiosyncracies like the outliers highlighted here, mean-sensitive, L2 normalized metrics (e.g. cosine similarity or Pearson correlation) will inevitably weigh the comparison of vectors along the highest-valued dimensions. In the case of positional artefacts propagating through the BERT and RoBERTa networks, the basis of comparison is inevitably steered towards whatever information is captured in those dimensions. Furthermore, since such outlier values show little variance across vectors, proxy metrics of anisotropy like measuring the average cosine similarity across random words (detailed in §\ref{sec:geo}) will inevitably return an exceedingly high similarity, no matter what the context. When cosine similarity is viewed primarily as means of semantic comparison between word or sentence vectors, the prospect of calculating cosine similarity for a benchmark like WiC or STS-B becomes erroneous. Though an examination of distance metrics is outside the scope of this study, we acknowledge similar points as having been addressed in regards to static word embeddings \citep{mimno2017strange} as well as contextualized ones \citep{li-etal-2020-sentence}. Likewise, we would like to stress that our manual clipping operation was performed for illustrative purposes and that interested researchers should employ more systematic post-hoc normalization strategies, e.g. whitening \citep{su2021whitening}, when working with hidden states directly. 

Relatedly, the anisotropic nature of the vector space that persists even after clipping the outliers suggests that positional artefacts are simply part of the explanation. Per this point, \citet{gao2019representation} prove that, in training any sort of model with likelihood loss, the representations learned for tokens being predicted will be naturally be pushed away from most other tokens in order to achieve a higher likelihood. They relate this observation to the Zipfian nature of word distributions, where the vast majority of words are infrequent. \citet{li2020sentence} extend this insight specifically to BERT and show that, while high frequency words concentrate densely, low frequency words are much more sparsely distributed. Though we do not attempt to dispute these claims with our findings, we do hope our experiments will highlight the important role that positional embeddings play in the representational geometry of Transformer-based models. Indeed, recent work has demonstrated that employing relative positional embeddings and untying them from the simultaneously learned word embeddings has lead to impressive gains for BERT-based architectures across common benchmarks \cite{he2020deberta,ke2020rethinking}. It remains to be seen how such procedures affect the representations of such models, however. 

Beyond this, it is clear that LayerNorm is the reason positional artefacts propagate though model representations in the first place. Indeed, our experiments show that the outlier dimension observed for BERT is tied directly to the \texttt{[CLS]} token, which always occurs at the requisite 1st position ---- despite having no linguistic bearing on the sequence of observed tokens being modeled.
However, the fact that RoBERTa (which employs a similar delimiter) retains outliers originating from different positions' embeddings implies that the issue of artefact propagation is not simply a relic of task design. It is possible that whatever positional idiosyncrasies contribute to a task's loss are likewise retained in their respective embeddings. In the case of BERT, the outlier dimension may be granted a large negative weight in order to differentiate the (privileged) 1st position between all others. This information being reconstructed by the LayerNorm parameters, which are shared for all positions in the sequence length, and then propagated up through the Transformer network is a phenomenon worthy of further attention. 

\section{Related work}

In recent years, an explosion of work focused on understanding the inner workings of pretrained neural language models has emerged. One line of such work investigates the self-attention mechanism of Transformer-based models, aiming to e.g. characterize its patterns or decode syntactic structure \citep{ raganato-tiedemann-2018-analysis,vig2019visualizing, marecek-rosa-2018-extracting, voita-etal-2019-analyzing, clark-etal-2019-bert, kobayashi-etal-2020-attention}. Another line of work analyzes models' internal representations using probes. These are often linear classifiers that take representations as input and are trained with supervised tasks in mind, e.g. POS-tagging, dependency parsing \citep{tenney2018what,liu-etal-2019-linguistic,lin2019open,hewitt-manning-2019-structural,zhao2020quantifying}. In such work, high probing accuracies are often likened to a particular model having ``learned'' the task in question. 


Most similar to our work, \citet{ethayarajh-2019-contextual} investigate the extent of ``contextualization'' in models like BERT, ELMo, and GPT-2 \citep{radford2019language}. Mainly, they demonstrate that the contextualized vectors of all words are non-isotropic across all models and layers. However, they do not indicate why these models have such properties. Also relevant are the studies of \citet{dalvi2018grain}, who introduce a neuron-level analysis method, and \citet{durrani-etal-2020-analyzing}, who use this method to analyze individual neurons in contextualized word vectors. Similarly to our experiment, \citet{durrani-etal-2020-analyzing} train a linear probe to predict linguistic information stored in a vector. They then employ the weights of the classifier as a proxy to select the most relevant neurons to a particular task. In a similar vein, \citet{coenen2019visualizing} demonstrate the existence of syntactic and semantic subspaces in BERT representations. 


\section{Conclusion}

In this paper, we called attention to sets of outlier neurons that appear in BERT and RoBERTa's internal representations, which bear consistently large values when compared to the distribution of values of all other neurons. In investigating the origin of these outliers, we employed a neuron-level analysis method which revealed that they are artefacts derived from positional embeddings and Layer Normalization. Furthermore, we found that outliers are a major cause for the anisotrophy of a model's vector space \citep{ethayarajh-2019-contextual}. Clipping them, consequently, can make the vector space more directionally uniform and increase the similarity between words' contextual representations. In addition, we showed that outliers can distort results when investigating word sense within contextualized representations as well as obtaining sentence embeddings via mean pooling, where removing them leads to uniformly better results. Lastly, we find that ``clipping'' does not affect models' performance on three supervised tasks.

It is important to note that the exact dimensions at which the outliers occur will vary pending different initializations and training procedures (as evidenced by our own RoBERTa model). As such, future work will aim at investigating strategies for mitigating the propagation of these artefacts when pretraining. Furthermore, given that both BERT and RoBERTa are masked language models, it will be interesting to investigate whether or not similar artefacts occur in e.g. autoregressive models like GPT-2 \citep{radford2019language} or XLNet \citep{NEURIPS2019_dc6a7e65}. Per the insights of \citet{gao2019representation}, it is very likely that the representational spaces of such models are anisotropic, but it is important to gauge the extent to which this can be traced to positional artefacts. 

\paragraph{Authors' Note}
We would like to mention \citet{kovaleva2021bert}'s contemporaneous work, which likewise draws attention to BERT's outlier neurons. While our discussion situates outliers in the context of positional embeddings and vector spaces, \citet{kovaleva2021bert} offer an exhaustive analysis of LayerNorm parameterization and its impact on masked language modeling and finetuning. We refer the interested reader to that work for a thorough discussion of LayerNorm's role in the outlier neuron phenomenon. 

\paragraph{Acknowledgments}
We would like to thank Joakim Nivre and Daniel Dakota for fruitful discussions and the anonymous reviewers for their excellent feedback.

\bibliography{eacl2021}
\bibliographystyle{acl_natbib}

\newpage
\appendix
\section{Outliers of distilled and large models}\label{app:outlier-distilled-large}

\begin{figure}
    \centering
    \includegraphics[width=5cm]{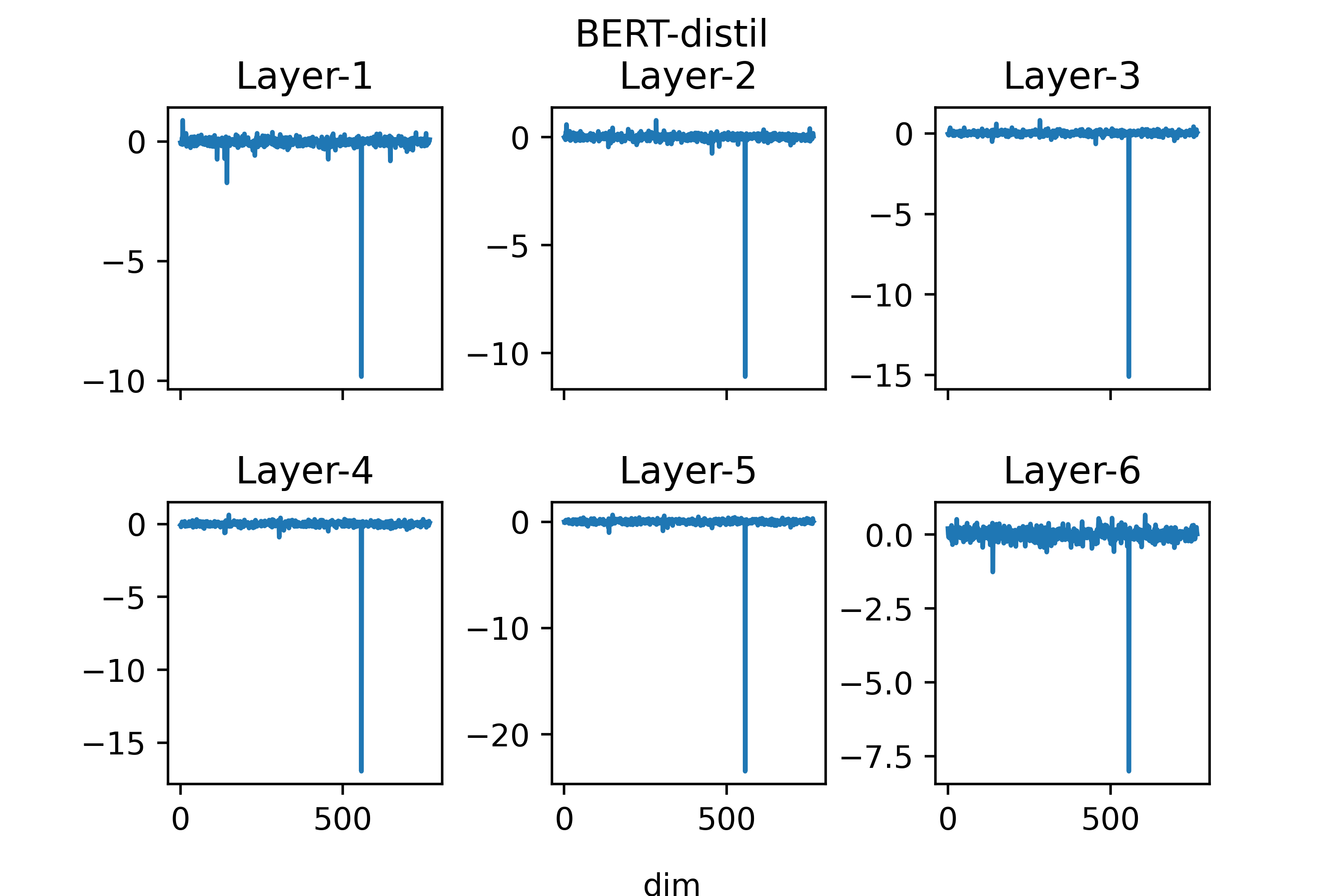}
    \caption{Average vectors for each layer of BERT-distil.}
    \label{fig:bert-distil}
\end{figure}

\begin{figure}
    \centering
    \includegraphics[width=5cm]{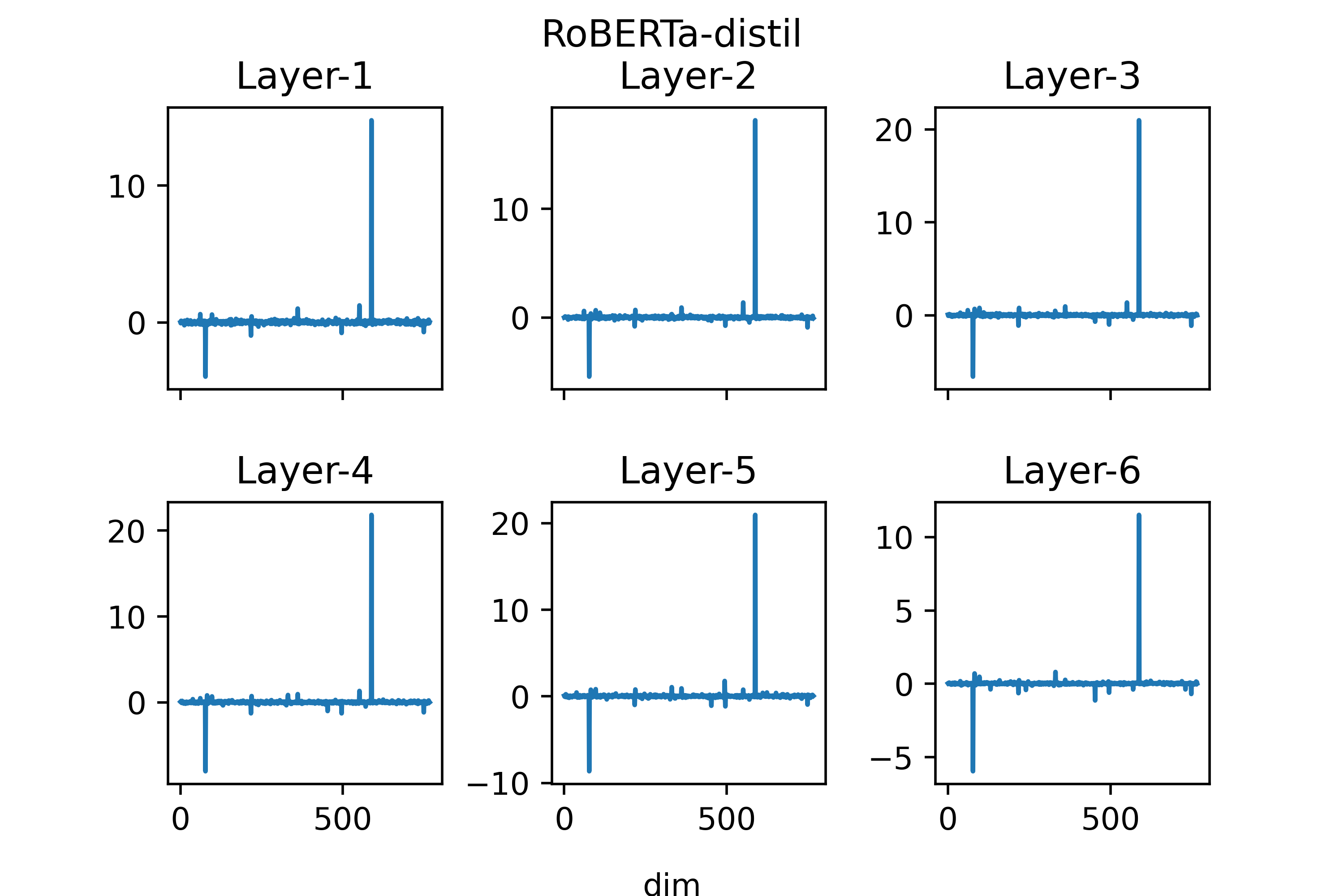}
    \caption{Average vectors for each layer of RoBERTa-distil.}
    \label{fig:Roberta-distil}
\end{figure}

\begin{figure}
    \centering
    \includegraphics[width=7.5cm]{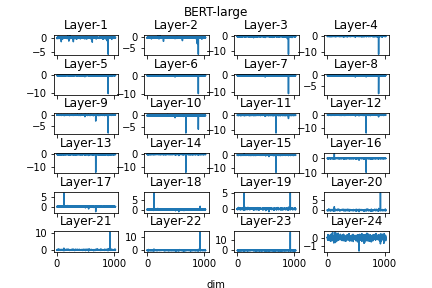}
    \caption{Average vectors for each layer of BERT-large.}
    \label{fig:bert-large}
\end{figure}

\begin{figure}
    \centering
    \includegraphics[width=7.5cm]{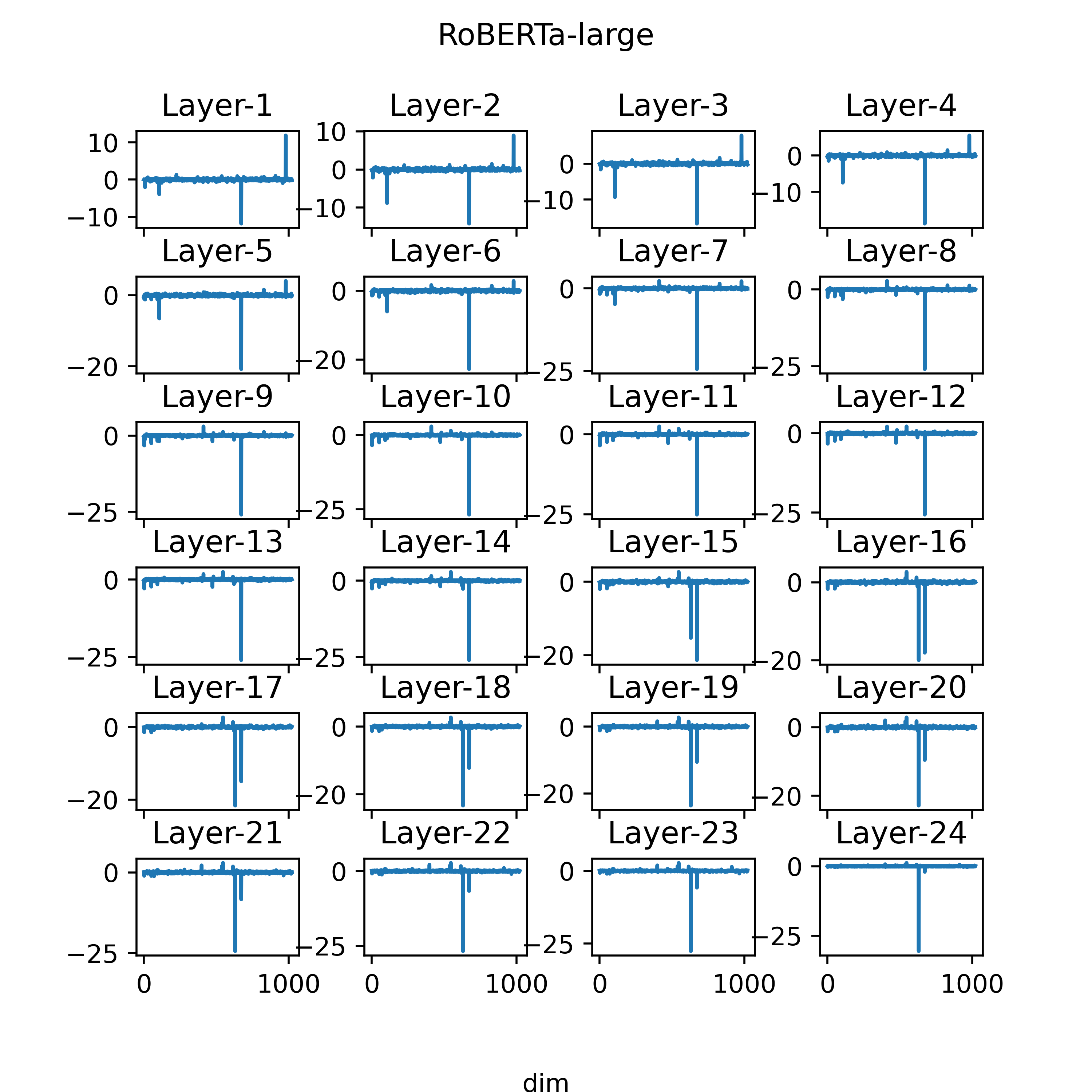}
    \caption{Average vectors for each layer of RoBERTa-large.}
    \label{fig:roberta-large}
\end{figure}

For BERT-distil, Figure \ref{fig:bert-distil} shows the patterns of BERT-distil across all layers. The $557^{th}$ element is an outlier. For RoBERTa-distil, Figure \ref{fig:Roberta-distil} shows the patterns of RoBERTa-distil across all layers. the $77^{th}$ and $588^{th}$ elements are two outliers. For BERT-large, Figure \ref{fig:bert-large} shows the patterns of BERT-large across all layers. From the first layer to the tenth layer, the $896^{th}$ element is an outlier. From the tenth layer to the seventeenth layer, the $678^{th}$ element is an outlier. From the sixteenth layer to the nineteenth layer, the $122^{nd}$ element is an outlier. From the nineteenth layer to the twenty-third layer, the $928^{th}$ element is an outlier. The final layer does not have outliers. For RoBERTa-large, Figure \ref{fig:roberta-large} shows the patterns of RoBERTa-large across all layers. From the first layer to the twenty-third layer, the $673^{rd}$ element is an outlier. From the fifteenth layer to the final layer, the $631^{st}$ element is an outlier. From the first layer to the sixth layer, the $981^{st}$ element is an outlier.

\section{Neuron-level analysis}

\subsection{Heatmaps of base models}\label{app:heatmap}

\begin{figure}
    \centering
    \includegraphics[height=4cm]{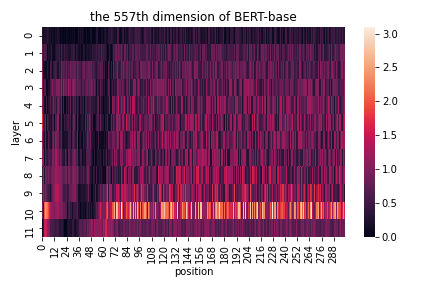}
    \includegraphics[height=4cm]{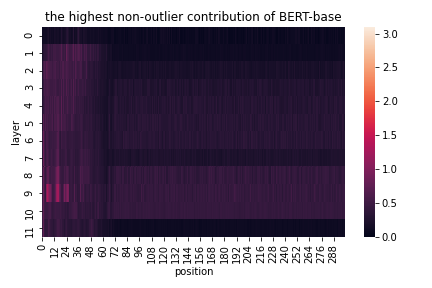}
    \caption{Up: contribution values heatmap of the outlier neuron of BERT-base. Down: the highest non-outlier contribution value of BERT-base.}
    \label{fig:bert-heatmap}
\end{figure}

\begin{figure}
    \centering
    \includegraphics[height=4cm]{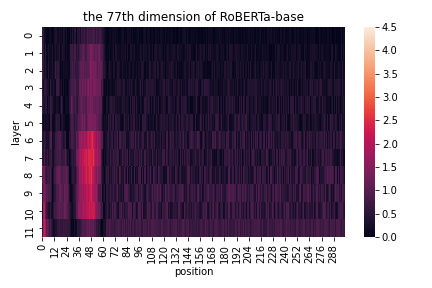}
    \includegraphics[height=4cm]{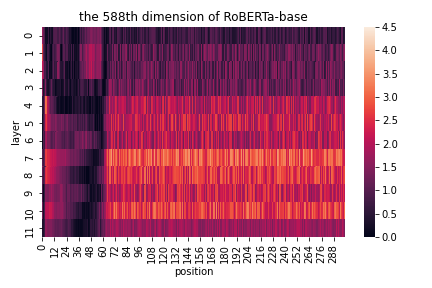}
    \includegraphics[height=4cm]{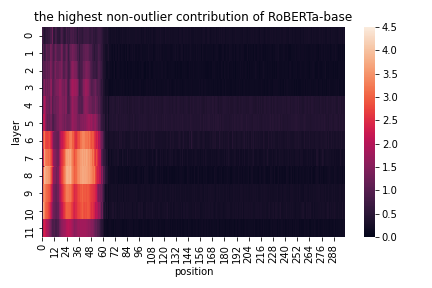}
    \caption{Up: contribution values heatmap of the 77th dimension of RoBERTa-base. Mid: contribution values heatmap of the 588th dimension of RoBERTa-base. Down: the highest non-outlier contribution value of RoBERTa-base.}
    \label{fig:RoBERTa-heatmap}
\end{figure}

Figure \ref{fig:bert-heatmap} and \ref{fig:RoBERTa-heatmap} show the heatmaps of the outlier neurons and the highest non-outlier contribution values.

\subsection{Distilled and large models}\label{app:analysis-distilled-large}

\begin{figure}
    \centering
    \includegraphics[height=4cm]{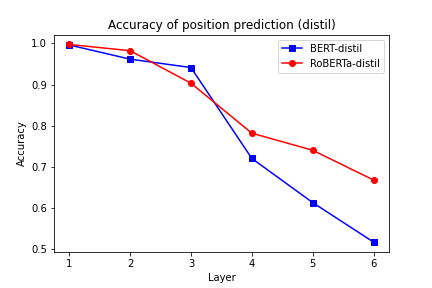}
    \includegraphics[height=4cm]{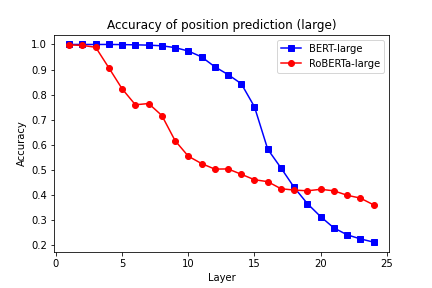}
    \caption{Up: accuracy of position prediction of distilled models. Down: accuracy of position prediction of large models.}
    \label{fig:acc-distil}
\end{figure}
Figure \ref{fig:acc-distil} show the accuracy scores of position prediction of distilled and large models.\\


\begin{figure}
    \centering
    \includegraphics[height=4cm]{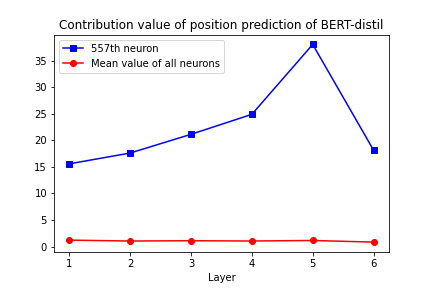}
    \includegraphics[height=4cm]{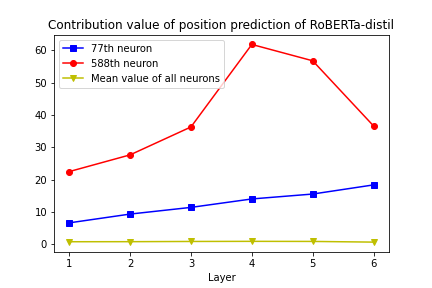}
    \caption{The contribution value of distilled models' outlier neurons on position prediction.}
    \label{fig:bert-distil-contri}
\end{figure}


\noindent\textbf{Distil-models}\indent Figure \ref{fig:bert-distil-contri} shows the contribution value of distilled models' outlier neurons on position prediction.\\

\begin{figure}
    \centering
    \includegraphics[height=4cm]{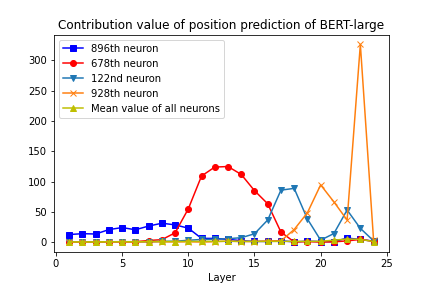}
    \includegraphics[height=4cm]{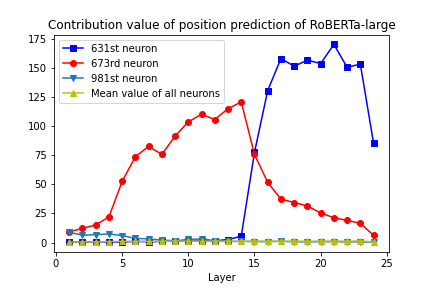}
    \caption{The contribution value of large models outlier neurons on position prediction.}
    \label{fig:bert-large-contri}
\end{figure}

\noindent\textbf{Large-models}\indent Figure \ref{fig:bert-large-contri} shows the contribution value of large models' outlier neurons on position prediction. 

\section{Our Pre-training Models}

\subsection{Hyper-parameters}\label{app:pre-train}

Table \ref{tab:encoder} shows the hyper-parameters of pre-training our RoBERTa-base models.

\begin{table}
    \centering
    \small
    \begin{tabular}{lc}
        \hline
        \textbf{Hyper-parameter} & Our RoBERTa-base \\
        \hline
        \hline
        Number of Layers & 12\\
        Hidden size & 768\\
        FNN inner hidden size & 3072\\
        Attention Heads & 12 \\
        Attention Head size & 64 \\
        Dropout & 0.1\\
        Warmup Steps & 10k\\
        Max Steps & 200k\\
        Learning Rates & 1e-4\\
        Batch Size & 256\\
        Weight Decay & 0.01\\
        Learning Rate Decay & Polynomial\\
        Adam ($\epsilon$, $\beta_1$, $\beta_2$) & (1e-6, 0.9, 0.98)\\
        Gradient Clipping & 0.5\\
        \hline
    \end{tabular}
    \caption{Hyper-parameters for pre-training our RoBERTa-base models.}
    \label{tab:encoder}
\end{table}

\subsection{Average subword vectors}\label{app:our-model}

\begin{figure}
    \centering
    \includegraphics[width=6.5cm]{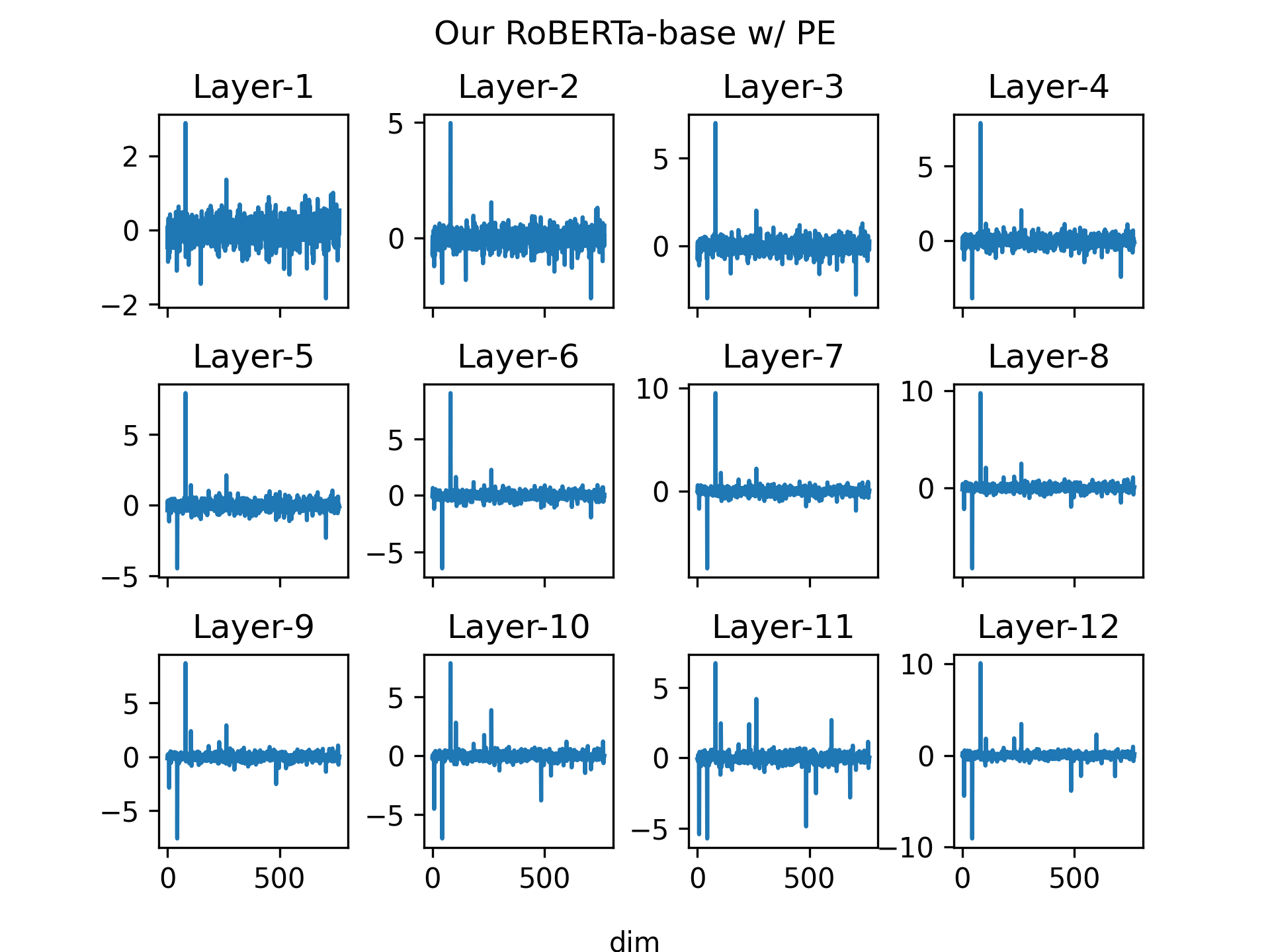}
    \includegraphics[width=6.5cm]{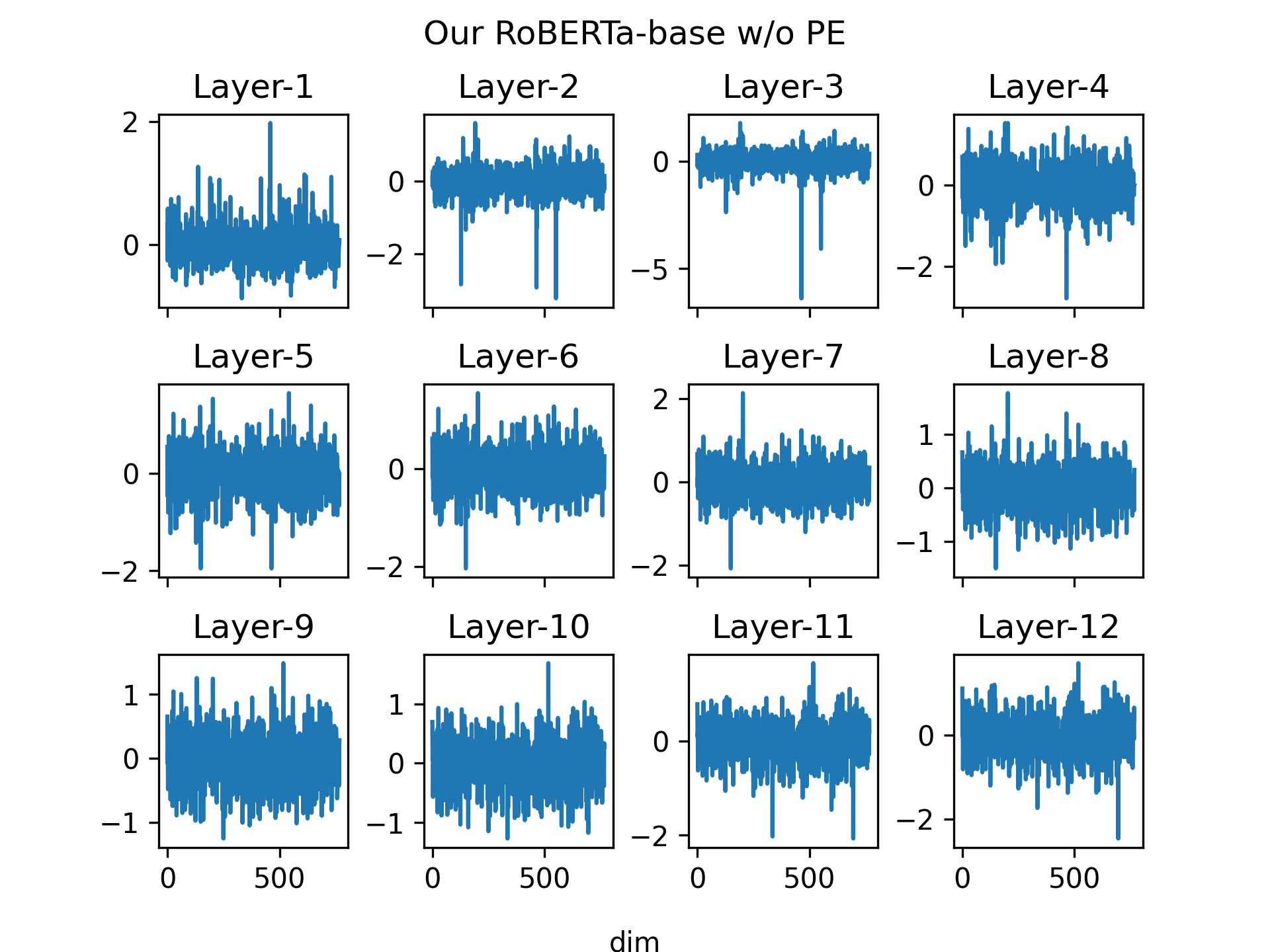}
    \caption{Average vectors for each layer of our RoBERTa-base w/ or w/o PE.}
    \label{fig:roberta-w-PE}
\end{figure}

Figure \ref{fig:roberta-w-PE} show the average vectors for each of our models.

\section{Clipping the outliers}\label{app:clip}

\subsection{Geometry of vector space}

\begin{figure}
    \centering
    \includegraphics[height=4cm]{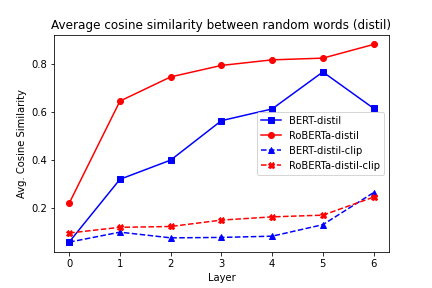}
    \includegraphics[height=4cm]{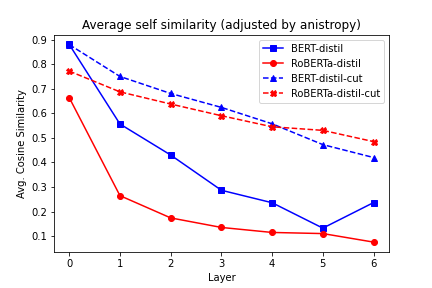}
    \caption{Up: average cosine similarity between random words of distil-models. Down: self-similarity measurement of BERT-distil and RoBERTa-distil (adjusted by anisotropy) before and after ``clipping the outliers''.}
    \label{fig:avg_distil}
\end{figure}

\noindent\textbf{Distil-models}\indent Figure \ref{fig:avg_distil} shows the anisotropic measurement of distilled models and the self-similarity measurement of distilled models.\\

\begin{figure}
    \centering
    \includegraphics[width=8.5cm]{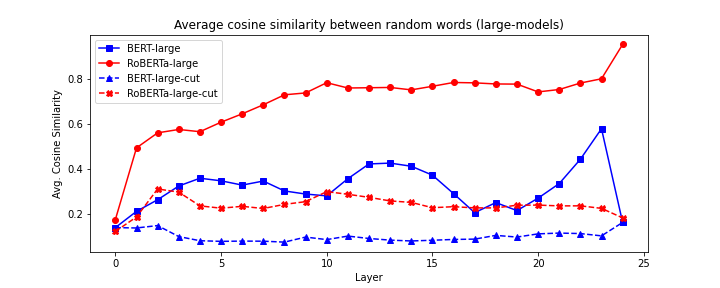}
    \caption{Average cosine similarity between random words of large-models.}
    \label{fig:avg-large}
\end{figure}

\begin{figure}
    \centering
    \includegraphics[width=8.5cm]{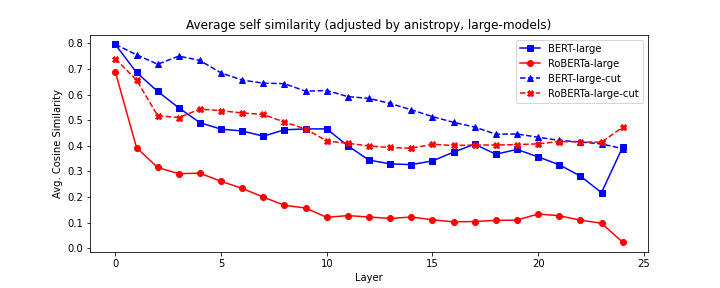}
    \caption{Self-similarity measurement of BERT-large and RoBERTa-large (adjusted by anisotropy) before and after ``clipping the outliers''.}
    \label{fig:self-large}
\end{figure}

\noindent\textbf{Large-models}\indent Figure \ref{fig:avg-large} shows the anisotropic measurement of large models and Figure  \ref{fig:self-large} shows the self-similarity measurement of large models. We ``clip'' different outlier neurons in different layers. For BERT-large, we zero-out the $896^{th}$ neuron from the first layer to the tenth layer, the $678^{th}$ neuron from the tenth layer to the seventeenth layer, the $122^{nd}$ neuron from the sixteenth layer to the nineteenth layer and the $928^{th}$ neuron from the nineteenth layer to the twenty-third layer. For RoBERTa-large, we zero-out the $673^{rd}$ neuron for all non-input layers, the $981^{st}$ neuron for the first 9 layers and the $631^{st}$ neuron for the last 10 layers.

\subsection{Word sense}\label{app:word-sense}


Table \ref{tab:wic-app} shows the accuracy scores of distill-models and large-models on WiC dataset before and after ``clipping the outliers''.

\begin{table}
    \centering
    \small
    \begin{tabular}{lccl}
        \hline
        Model & Layer & Threshold & Acc. \\
        \hline
        \hline
        \textbf{Baseline} & - & - & 50.0\%\\
        \hline
        \hline
        \textbf{Before clipping}\\
        BERT-distil & 5 & 0.9 & 66.5\%\\
        RoBERTa-distil & 5 & 0.9 & 63.7\%\\
        BERT-large & 12 & 0.7 & 70.2\%\\
        RoBERTa-large & 10 & 0.9 & 70.4\%\\
        \hline
        \hline
        \textbf{After clipping}\\
        BERT-distil-clip & 6 & 0.6 & \textbf{67.3\%}\\
        RoBERTa-distil-clip & 5 & 0.6 & \textbf{66.7\%}\\
        BERT-large-clip & 12 & 0.6 & \textbf{70.3\%}\\
        RoBERTa-large-clip & 16 & 0.6 & \textbf{71.3\%}\\
        \hline
    \end{tabular}
    \caption{The best accuracy scores on WiC dataset for distilled and large models. \textbf{Bold} indicates that the best result increases after clipping.}
    \label{tab:wic-app}
\end{table}

\subsection{Sentence embedding}

Table \ref{tab:sent-distil} shows the results on semantic textual similarity tasks of distilled models before and after ``clipping the outliers''.

\begin{table}
    \centering
    \scriptsize
    \begin{tabular}{l|ll|ll}
        \hline
        Dataset & \tabincell{c}{BERT\\distil} & \tabincell{c}{RoBERTa\\distil} & \tabincell{c}{BERT\\distil\\clip} & \tabincell{c}{RoBERTa\\distil\\clip}\\
        \hline
        \hline
        STS-B & 59.65(6) & 56.06(5) & 56.62(6) & \textbf{58.47(5)}\\
        SICK-R & 62.64(6) & 62.63(5) & 62.42(6) & \textbf{62.73(6)}\\
        STS-12 & 42.96(1) & 40.19(1) & \textbf{46.47(1)} & \textbf{42.36(1)}\\
        STS-13 & 59.33(1) & 56.42(5) & 55.74(1) & \textbf{60.64(6)}\\
        STS-14 & 53.81(6) & 49.59(6) & 50.57(1) & \textbf{52.51(2)}\\
        STS-15 & 61.40(6) & 65.10(5) & \textbf{61.48(1)} & \textbf{65.93(2)}\\
        STS-16 & 61.43(6) & 62.90(5) & 60.75(6) & \textbf{64.49(5)}\\
        \hline
    \end{tabular}
    \caption{Experimental results on semantic textual similarity of distilled models.  The number in the parenthesis denotes that this result belongs to the specific layer. \textbf{Bold} indicates that the best result increases after clipping.}
    \label{tab:sent-distil}
\end{table}

Table \ref{tab:sent-large} shows the results on semantic textual similarity tasks of large models before and after ``clipping the outliers''.

\begin{table}
    \centering
    \scriptsize
    \begin{tabular}{l|ll|ll}
        \hline
        Dataset & \tabincell{c}{BERT\\large} & \tabincell{c}{RoBERTa\\large} & \tabincell{c}{BERT\\large\\clip} & \tabincell{c}{RoBERTa\\large\\clip}\\
        \hline
        \hline
        STS-B & 62.56(1) & 59.71(19) & \textbf{66.43(3)} & \textbf{62.01(23)}\\
        SICK-R & 64.47(24) & 63.08(14) & \textbf{65.72(23)} & \textbf{63.50(16)}\\
        STS-12 & 54.05(1) & 44.72(1) & \textbf{56.44(3)} & \textbf{49.69(1)}\\
        STS-13 & 68.80(2) & 61.68(8) & \textbf{71.07(2)} & \textbf{62.82(10)}\\
        STS-14 & 60.46(1) & 51.39(8) & \textbf{63.35(1)} & \textbf{57.33(1)}\\
        STS-15 & 73.91(1) & 65.98(7) & \textbf{76.51(1)} & \textbf{69.71(1)}\\
        STS-16 & 66.35(17) & 66.50(14) & \textbf{71.41(3)} & \textbf{68.25(11)}\\
        \hline
    \end{tabular}
    \caption{Experimental results on semantic textual similarity of large models. The number in the parenthesis denotes that this result belongs to the specific layer. \textbf{Bold} indicates that the best result increases after clipping.}
    \label{tab:sent-large}
\end{table}

\end{document}